\documentclass[conference]{IEEEtran}
\IEEEoverridecommandlockouts
\usepackage{cite}
\usepackage{amsmath,amssymb,amsfonts}
\usepackage{algorithm}
\usepackage{algpseudocode}
\usepackage{graphicx}
\usepackage{textcomp}
\usepackage{xcolor}
\usepackage{booktabs}
\usepackage{multirow}
\usepackage{adjustbox}
\usepackage{float}
\usepackage{placeins}
\usepackage{microtype}
\usepackage[hidelinks]{hyperref}
\usepackage{orcidlink}

\def\BibTeX{{\rm B\kern-.05em{\sc i\kern-.025em b}\kern-.08em
    T\kern-.1667em\lower.7ex\hbox{E}\kern-.125emX}}

\begin{document}

\title{BrainFocus: EEG-Guided ROI Selection for Efficient Vision-Language Models}

\author{\IEEEauthorblockN{Yihui Peng*\orcidlink{0009-0003-6845-4826}, Guorui Lu*\orcidlink{0009-0005-3455-0670},~\IEEEmembership{Student Member,~IEEE}, Qinyu Chen\orcidlink{0009-0005-9480-6164},~\IEEEmembership{Member,~IEEE}}
\thanks{*Yihui Peng and Guorui Lu equally contributed to the work.}
\thanks{Yihui Peng, Guorui Lu and Qinyu Chen are with Leiden Institute of Advanced Computer Science (LIACS), Leiden University, The Netherlands. (\{g.lu, q.chen\}@liacs.leidenuniv.nl)}
\thanks{Corresponding code and data: https://github.com/Yihui-Peng/BrainFocus}
}
% and \href{https://ieee-dataport.org/documents/brainfocus-dataset-ai-generated-cluttered-scene-images-bounding-boxes-and-vqa-metadata}{IEEE DataPort}.

\maketitle

\begin{abstract}
Vision-language models (VLMs) achieve strong visual question answering (VQA) performance, but processing large cluttered images is computationally expensive when only a small region is relevant. 
Electroencephalography (EEG) signals, which capture human neural responses to visual stimuli, can provide a human-derived semantic cue about the region of interest (ROI).
However, EEG-guided visual category decoding remains imperfect, making direct ROI routing unreliable.
% However, EEG-guided visual category decoding remains imperfect, particularly as the number of candidate categories increases, making direct ROI routing unreliable.
In this work, we propose BrainFocus, a reliable EEG-guided efficient VLM framework for VQA. 
An EEG classifier predicts a target category, and a YOLO detector localizes the matching ROI.
The VLM receives the cropped ROI only when both predictions pass confidence thresholds; otherwise, it processes the full image. 
For evaluation, we build on EEG-ImageNet to construct a 40-class benchmark comprising generated cluttered images and real object-centric images, with target-ROI annotations and 600 English visual question–answer pairs. 
Across Qwen3.5-VL 2B, 4B, and 9B models, BrainFocus improves VQA accuracy by 4.14--9.87 percentage points (pp) on cluttered scenes while reducing input tokens and total tokens by 23.2\%--39.4\% and 23.2\%--39.3\%, and end-to-end floating-point operations (FLOPs) by 23.2\%--39.5\%. 
% On object-centric images, accuracy changes by only -0.40 to +0.51 pp while FLOPs decrease by 3.1--6.5\%. 
% These results demonstrate that our framework is more effective in complex scenarios, as cluttered scenes more closely reflect real-world usage. This also confirms that our framework is more suitable for real-life scenarios. 
These results demonstrate that EEG can guide efficient VLM inference even when its semantic decoding is imperfect.
\end{abstract}


\section{Introduction} \label{introduction}

Vision-language models (VLMs) support strong visual question answering (VQA) and general multimodal reasoning \cite{antol2015vqa, li2023blip2, liu2023llava}. However, processing high-resolution images incurs substantial cost~\cite{dosovitskiy2021vit}.
In cluttered scenes, answering a question may require information from only a small target region, while much of the image contributes little to the answer.
%, even though the answer may depend on only one object. 
Existing approaches improve efficiency mainly through modifying the visual encoder, visual token compression, or pruning inside the model \cite{chen2024fastv,zhang2025sparsevlm,yang2025visionzip,vasu2025fastvlm,hu2024mqt}. 
A complementary approach is to select a relevant region of interest (ROI) before VLM inference, reducing the visual input without modifying the VLM architecture. This approach requires a reliable cue about which region to retain.
% An external region-of-interest (ROI) route is model-agnostic: localize the question-relevant object first, then submit only its crop. Its practical value depends on obtaining a reliable cue about which object matters.
 
% Gaze is another common cue of human attention, but eye position does not always align with the area of attention and may therefore misidentify the ROI \cite{posner1980orienting}. Nevertheless, EEG can always reflect what people see with their eyes and what they think in their minds.

Electroencephalography (EEG) provides a human-derived semantic cue, as visually evoked responses contain decodable object-category information~\cite{spampinato2017deep,palazzo2021decoding,gifford2022large,grootswagers2022things,zhu2024eegimagenet}. Although EEG does not directly provide spatial coordinates, its decoded category can condition an object detector to localize the target ROI, linking neural decoding to efficient VLM inference. However, EEG category decoding remains imperfect, with reported accuracy around 60\% for broad category spaces~\cite{zhu2024eegimagenet}. Incorrect predictions may direct the detector toward irrelevant objects and remove evidence needed for VQA, motivating a selective mechanism that crops only reliable ROIs.

Existing EEG-based visual studies mainly investigate classification, decoding, representation analysis, or reconstruction \cite{spampinato2017deep,palazzo2021decoding,gifford2022large,grootswagers2022things,zhu2024eegimagenet}. Meanwhile, efficient VLM methods mainly exploit information from images, text queries, or internal model representations to reduce visual computation~\cite{cao2023pumer,chen2024fastv,zhang2025sparsevlm,yang2025visionzip}. Region-based methods such as CROP~\cite{guo2025crop} employ a separate lightweight VLM to identify a relevant region. 
However, these directions remain largely separate, and the use of noisy EEG-derived semantic cues for pre-inference ROI selection without modifying the VLM backbone remains underexplored.

% 流程图
\begin{figure*}[!t]
    \centering
    \includegraphics[width=\textwidth]{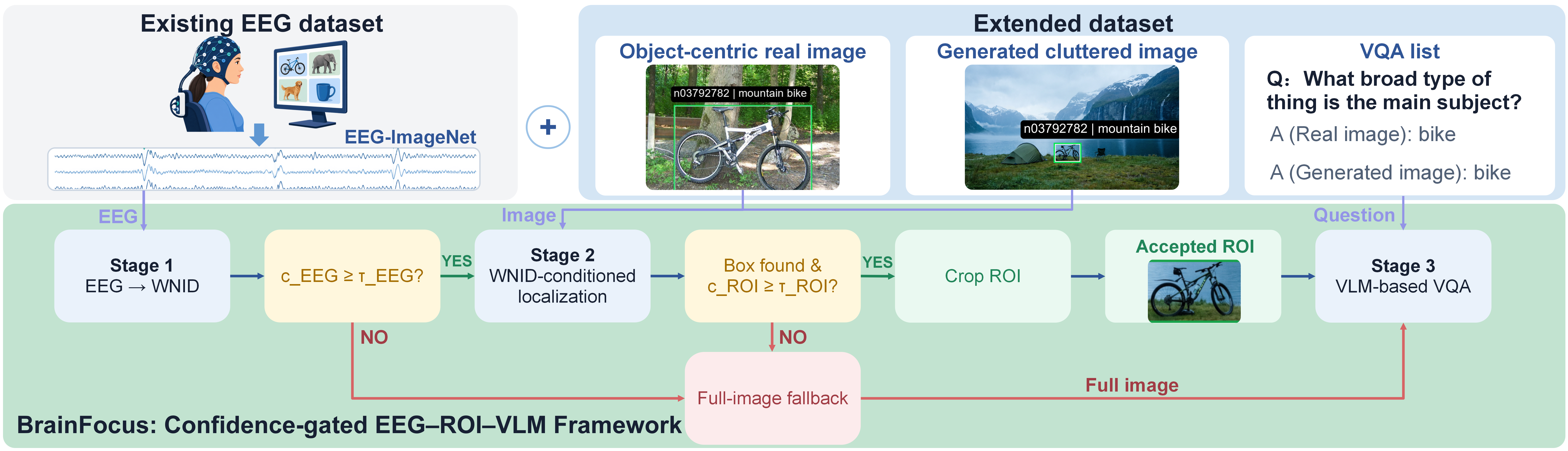}
    \caption{Overview of the proposed BrainFocus EEG-ROI-VLM framework.}
    \label{fig:pipeline}
    \vspace{-2mm}
\end{figure*}

We bridge this gap with \textbf{BrainFocus}, a reliable EEG-guided efficient VLM inference framework for VQA.
% An EEG classifier predicts a target category, and a category-conditioned detector localizes the matching object. Two sequential confidence gates determine whether the resulting ROI is passed to the VLM. Rejection for unsuccessful ROI attempts invokes a full-image fallback. 
Our work makes the following contributions: 
\begin{itemize}
    \item We connect EEG category decoding to ROI selection for efficient VLM inference, without modifying the VLM backbone or introducing an auxiliary VLM.
    \item We introduce a dual-confidence gating strategy with full-image fallback to accommodate uncertainty in EEG decoding and object localization. The strategy selectively accepts crops, enabling imperfect EEG-derived semantic cues to support efficient VLM inference.
    \item We construct a category-matched evaluation benchmark by extending the EEG-ImageNet dataset~\cite{zhu2024eegimagenet} with generated cluttered images, real object-centric images, ROI labels, and VQA pairs on 15 subjects. BrainFocus improves VQA accuracy by 4.14--9.87 percentage points (pp) while reducing end-to-end FLOPs by 23.2\%--39.5\% on generated cluttered scenes with Qwen3.5-VL-2B, 4B, and 9B~\cite{qwen2026qwen35}.
\end{itemize}

BrainFocus demonstrates that EEG can guide efficient VLM inference despite imperfect EEG predictions, improving VQA accuracy while reducing computation on generated cluttered scenes, which can be useful for applications like VR/AR.

% \begin{itemize}
%     \item A new dataset with high-resolution cluttered LLM-generated images, real-centric images, and a corresponding 40-class VQA list. It is used to verify the effectiveness of the framework in both target-centric scenarios and cluttered scenarios.
%     \item We formulate EEG as an imperfect semantic routing signal rather than a direct spatial predictor, connecting neural decoding to model-agnostic visual input reduction without introducing an additional VLM.
%     \item We introduce a confidence and full-image fallback strategy calibrated for accepted-crop reliability to avoid the negative influence of unreliable ROI crops.
%     \item We evaluate the complete framework across 4 class settings, 2 image domains, 15 subjects, and 3 VLM sizes using accuracy, visual tokens, and floating-point operations (FLOPs).
% \end{itemize}

% Despite EEG visual stimuli classification accuracy as low as around 50\% \cite{zhu2024eegimagenet}, our BrainFocus framework can still improve VQA accuracy and reduce computation in cluttered scenes. Such high-resolution complex scenes are closer to practical application scenarios. The key insight is that EEG doesn't have to identify every target accurately. With an appropriate strategy, imperfect semantic signals can still benefit the VLM inference. 
% It only needs to select a sufficiently reliable subset of trials on which ROI routing is beneficial, while the remaining trials are protected by the fallback.

\section{Dataset}
\label{sec:dataset}

We build on EEG-ImageNet~\cite{zhu2024eegimagenet} to evaluate the BrainFocus framework.
The expanded dataset combines existing EEG recordings with two image domains (i.e. real object-centric images and generated cluttered images), ROI labels, and a VQA pair set.
The two domains allow us to examine how the benefit of ROI selection varies with scene composition and target size. Table~\ref{tab:vqa_logical_trials} summarizes the data used at each stage.

\subsection{EEG-ImageNet Dataset}
\label{subsec:dataset_eeg_settings}

EEG-ImageNet contains recordings from 16 participants viewing 4000 ImageNet-21k stimuli across 40 coarse and 40 fine categories, with 50 images per category \cite{zhu2024eegimagenet,deng2009imagenet}. Recordings used 62 electrodes at 1000 Hz during each 0.5-s presentation. Each trial links an EEG trace to its image index and ImageNet WordNet ID (WNID). In this paper, we conduct all experiments on coarse categories (e.g., mountain bike, panda, and banana) because they cover a much wider range of categories than fine categories (e.g., strawberry, watermelon, and grape). We excluded Subject 12 due to missing data, so we ultimately conducted the experiment on 15 subjects.

\subsection{Real and Generated Image Domains}
\label{subsec:dataset_image_domains}

% 真实集
The real-image-test domain contains 600 photographs from 40 classes, with 15 images per class and one validated, non-empty, in-bounds target box per image. Most targets are large, centered, and isolated, limiting their representation of complex scenes (shown as the extended dataset part in Fig.~\ref{fig:pipeline}).
% 生成集
To complement this domain, we generate 50 realistic-style cluttered images per class using MiniMax Image-01 and ByteDance Seedream 4.0 through their official APIs, yielding 2000 images with broader backgrounds, distractors, and multi-object layouts. We designed the prompts to instruct generative models to generate realistic scenes in which the target object co-occurs with diverse visual elements.
This generation method can handle rare natural co-occurrence situations (such as the co-occurrence of giant pandas and German shepherds), and can also be used to test whether our framework can identify the relevant ROI in complex realistic scenes.
In both image domains, EEG trials and images are paired at the category level rather than at the individual-image level.

\subsection{VQA List Construction}
\label{subsec:dataset_vqa}

% 表格1 数据统计表
% \begin{table}[!t]
%     \centering
%     \caption{Dataset details under different settings. \textcolor{red}{table notes}}
%     \label{tab:vqa_logical_trials}
%     \begin{adjustbox}{max width=\linewidth}
%     \begin{tabular}{lcccccc}
%         \toprule
%         Setting & Generated img & Real img & EEG & Q\&A  &  Subjects & Trials \\
%         \midrule
%         10-class & 300/50/150   & 150 & 500 & 150 & 15 & 2250 \\
%         20-class & 600/100/300  & 300 & 1000 & 300 & 15 & 4500 \\
%         30-class & 900/150/450  & 450 & 1500 & 450 & 15 & 6750 \\
%         40-class & 1200/200/600 & 600 & 2000 & 600 & 15 & 9000 \\
%         \bottomrule
%     \end{tabular}
%     \end{adjustbox}
% \end{table}

\begin{table}[!t]
    \centering
    \caption{Dataset details under different settings.}
    \label{tab:vqa_logical_trials}
    \begin{adjustbox}{max width=\linewidth}
    \begin{tabular}{lccccc}
        \toprule
        \multirow{3}{*}{\shortstack{Setting\\($C$-class)}}
        & \multicolumn{1}{c}{Stage 1}
        & \multicolumn{2}{c}{Stage 2}
        & \multicolumn{1}{c}{Stage 3}
        & \multirow{3}{*}{Valid subj.} \\

        \cmidrule(lr){2-2}
        \cmidrule(lr){3-4}
        \cmidrule(lr){5-5}

        & EEG trials per subj.
        & Generated img.
        & Real img.
        & Q\&A pairs
        & \\

        & (Train/Val/Test)
        & (Train/Val/Test)
        & (Test)
        & (Test)
        & \\
        \midrule

        10-class & 300/50/150   & 300/50/150   & 150 & 150 & 15 \\
        20-class & 600/100/300  & 600/100/300  & 300 & 300 & 15 \\
        30-class & 900/150/450  & 900/150/450  & 450 & 450 & 15 \\
        40-class & 1200/200/600 & 1200/200/600 & 600 & 600 & 15 \\

        \bottomrule
    \end{tabular}
    \end{adjustbox}
    \vspace{-9pt}
    % \begin{minipage}{\linewidth}
    %     \scriptsize
    %     \textit{Note:} For the $C$-class setting, in stage 2, the generated
    %     images have $30C$, $5C$, and $15C$ samples for training, validation, and testing. The real images have $15C$ testing samples. In stage 3, $15C$ VQA pairs are included.
    % \end{minipage}
\end{table}

% For a $C$-class setting, EEG and generated-image counts are $C\!\times\!(30/5/15)$ for training/validation/test. Real images and Q\&A items are test-only, each with $15C$ entries. One trial is one subject's EEG paired with one test image and its question (Stage~3). Thus, Trials $=15\text{ subjects}\times C\text{ classes}\times15\text{ test slots}$ per image domain and VLM.

We construct a 600-item English VQA list for the 40 classes by pairing each test image per class with one question-answer item. 12 question types cover identity, category, partial attributes, living status, broad category, visual evidence, color, function, mobility, technology type, the general environment, and background where the target object is likely to appear. The list contains 320 yes/no (53.3\%) and 280 short-phrase (46.7\%) answers. VQA examples in both domains are shown in Fig. \ref{fig:pipeline}.

\section{Methods}\label{sec:methods}
BrainFocus converts imperfect EEG predictions into safe input routing: reliable EEG-guided ROIs are sent to the VLM, whereas uncertain requests retain the full image to maintain answer quality.

\subsection{Three-Stage EEG--ROI--VLM Framework}
\label{subsec:method_framework}

\begin{algorithm}[!ht]
\caption{BrainFocus: confidence-gated VLM framework}
\label{alg:eeg_roi_vqa}
\begin{algorithmic}[1]
\Require EEG $x_{\mathrm{eeg}}$, image $I$, question $q$, thresholds $\tau_{\mathrm{eeg}}$/$\tau_{\mathrm{roi}}$
\Ensure Visual answer $\hat{a}$

\Statex\hspace*{-\algorithmicindent}\parbox[t]{0.98\linewidth}{
\textbf{Notation:}
\begin{list}{}{
    \setlength{\leftmargin}{1.5em}
    \setlength{\itemsep}{0pt}
    \setlength{\parsep}{0pt}
    \setlength{\topsep}{2pt}
}
    \item $\hat{w}$: predicted WNID;
    \item $\hat{b}$: predicted bounding box;
    \item $c_{\mathrm{eeg}}$ \& $c_{\mathrm{roi}}$: confidence scores of Stage 1 and 2;
    \item $I^\star$: selected visual input, either the full image or ROI.
\end{list}
}

\Statex
\State $I^\star \gets I$ %\Comment{Default full-image fallback}
\State $(\hat{w}, c_{\mathrm{eeg}}) \gets f_{\mathrm{EEG}}(x_{\mathrm{eeg}})$ %\Comment{Stage 1: EEG-to-WNID classification}

\If{$c_{\mathrm{eeg}} \geq \tau_{\mathrm{eeg}}$}
    \State $(\hat{b}, c_{\mathrm{roi}}) \gets f_{\mathrm{ROI}}(I, \hat{w})$ %\Comment{Stage 2: WNID-conditioned ROI localization}
    \If{$\hat{b}$ exists \textbf{and} $c_{\mathrm{roi}} \geq \tau_{\mathrm{roi}}$}
        \State $I^\star \gets \mathrm{Crop}(I, \hat{b})$ %\Comment{Use cropped ROI only when both stages are confident}
    \EndIf
\EndIf

\State $\hat{a} \gets \mathrm{VLM}(I^\star, q)$ %\Comment{Stage 3: VQA}
\State \Return $\hat{a}$
\end{algorithmic}
\end{algorithm}

Fig.~\ref{fig:pipeline} and Algorithm~\ref{alg:eeg_roi_vqa} summarize the three-stage framework. Stage 1 reproduces the EEG-ImageNet classifier \cite{zhu2024eegimagenet}. Differential-entropy features from the delta, theta, alpha, beta, and gamma bands are fed to a two-layer MLP with 256 and 128 hidden units \cite{duan2013differential}. The network predicts a top-1 WNID from the viewed stimulus. Stage 2 receives the image and the predicted WNID. A YOLO detector fine-tuned on the generated cluttered images returns candidate boxes and class scores \cite{redmon2016yolo,wang2023yolov7}. We select the highest-confidence box matching the predicted WNID. If none exists, no crop is returned. Stage 3 sends either this crop or the full image to VLM with the same question.

\subsection{Confidence-Gated Fallback Strategy}
\label{subsec:method_framework_strategy}

Because EEG category predictions and detector outputs are imperfect, unconditional cropping can result in a poor cropping effect, and excessive incorrect cropping will remove the necessary visual evidence needed for VQA. We therefore employ a sequential confidence-gated fallback strategy, which includes two gates: EEG gate and ROI gate. The EEG gate will first evaluate each request. If $c_{\mathrm{eeg}}<\tau_{\mathrm{eeg}}$, Stage~2 is skipped and the full image is sent directly to Stage~3. Otherwise, Stage~2 searches for a detection matching the predicted WNID. The cropped ROI is fed to Stage~3 only when a matching box exists and $c_{\mathrm{roi}}\geq\tau_{\mathrm{roi}}$, while all other requests use the full-image fallback. Thus, every request can reach Stage~3, whereas crop routing requires both confidence gates to pass.

For Stage~1, let $\mathbf{z}$ denote the EEG classifier logits. Following the widely used maximum-class-probability baseline for failure prediction \cite{hendrycks2017baseline,corbiere2019addressing}, we define $\hat{w}=\arg\max_k\operatorname{softmax}(\mathbf{z})_k$ as the predicted WNID, and its confidence is the corresponding maximum softmax score \cite{guo2017calibration}, $c_{\mathrm{eeg}}=\max_k\operatorname{softmax}(\mathbf{z})_k$. For Stage~2, we retain detections whose class matches $\hat{w}$, and use $c_{\mathrm{roi}}$, the detector-native confidence that YOLO outputs for the highest-scoring matching box. No matching box causes fallback. During calibration, an \emph{accepted} trial has a matching box and passes both gates, whereas it is \emph{reliable} only if $\hat{w}$ is correct and the selected box has target Intersection over Union (IoU) $\geq0.5$. \emph{Precision} is Reliable/Accepted, and the non-fallback rate is Accepted/7,500. We test threshold pairs on the held-out trials and select the pair with the highest non-fallback rate subject to Precision $\geq0.97$. This yields $(\tau_{\mathrm{eeg}},\tau_{\mathrm{roi}})=(0.20,0.05)$, balancing crop reliability against crop use.

% \begin{table}[H]
%     \centering
%     \caption{Confidence-Gate calibration on 7,500 validation trials.}
%     \label{tab:confidence_gate_calibration}
%     \scriptsize
%     \setlength{\tabcolsep}{2.5pt}
%     \renewcommand{\arraystretch}{1.05}
%     \resizebox{\columnwidth}{!}{%
%     \begin{tabular}{lcccc}
%         \toprule
%         Policy & $\tau_{\mathrm{eeg}}$ & $\tau_{\mathrm{roi}}$ & Precision (Reliable/Accepted) & Non-fallback \\
%         \midrule
%         Ungated & 0.00 & 0.00 & 0.9518 (3886/4083) & 0.5444 \\
%         $P\geq0.96$ & 0.05 & 0.05 & 0.9660 (3210/3323) & 0.4431 \\
%         \textbf{$P\geq0.97$} & \textbf{0.20} & \textbf{0.05} & \textbf{0.9703 (2842/2929)} & \textbf{0.3905} \\
%         $P\geq0.98$ & 0.15 & 0.30 & 0.9808 (1888/1925) & 0.2567 \\
%         \bottomrule
%     \end{tabular}}
% \end{table}

\begin{table}[t]
    \centering
    \caption{Confidence-Gate calibration on 7,500 validation trials.}
    \label{tab:confidence_gate_calibration}
    \scriptsize
    \setlength{\tabcolsep}{2.5pt}
    \renewcommand{\arraystretch}{1.05}
    \resizebox{\columnwidth}{!}{%
    \begin{tabular}{ccccc}
        \toprule
        Policy
        & $\tau_{\mathrm{eeg}}$
        & $\tau_{\mathrm{roi}}$
        & Precision (Reliable/Accepted)
        & Non-fallback \\
        \midrule

        Ungated
        & 0.00
        & 0.00
        & 0.9518 (3886/4083)
        & 0.5444 \\

        $P\geq0.96$
        & 0.05
        & 0.05
        & 0.9660 (3210/3323)
        & 0.4431 \\

        {\bfseries\boldmath $P\geq0.97$}
        & \textbf{0.20}
        & \textbf{0.05}
        & \textbf{0.9703 (2842/2929)}
        & \textbf{0.3905} \\

        $P\geq0.98$
        & 0.15
        & 0.30
        & 0.9808 (1888/1925)
        & 0.2567 \\
        \bottomrule
    \end{tabular}}
    \vspace{-3pt}
\end{table}

Even without confidence thresholds, YOLO does not always return a box matching the predicted WNID. As shown in Table~\ref {tab:confidence_gate_calibration}, the ungated non-fallback rate is only 0.5444. Gating lowers this rate to 0.3905 by rejecting 1,154 crops, including 110 accepted but unreliable ones, and raises precision from 0.9518 to 0.9703. This trade-off shows that the ungated crop set contains an unreliable subset that needed to be rejected.

\subsection{Semantic Answer Scoring}
\label{subsec:method_semantic_scoring}

VQA answers are scored using a two-level protocol. We first use exact matching or recognized synonym matching to score the simple VQA outputs such as yes/no answers. Then, short-phrase answers are submitted to GPT-5.5~\cite{openai2026gpt55} for further scoring. The question, expected answer, candidate answers, answer format, and evaluation description are attached simultaneously. The evaluators are required to provide the evaluation results along with a brief reason for double-checking.

\section{Experimental Results}
\label{sec:experiment results}
We evaluate whether the BrainFocus framework reduces VLM computation without sacrificing answer quality, and how the trade-off changes across image domains, subjects, class counts, and VLM model sizes.

\begin{table}[!t]
    \centering
    \setlength{\belowcaptionskip}{2pt}
    \caption{Stage-wise performance and routing results (\%).}
    \label{tab:stagewise_results}

    \scriptsize
    \setlength{\tabcolsep}{1.8pt}
    \renewcommand{\arraystretch}{0.95}

    \begin{adjustbox}{max width=\columnwidth}
    \begin{tabular}{@{}c cc ccc ccc@{}}
        \toprule
        \multirow{2}{*}{$C$}
        & \multicolumn{2}{c}{Stage 1: EEG}
        & \multicolumn{3}{c}{Stage 2: Detector}
        & \multicolumn{3}{c}{Test routing (generated/real)} \\
        \cmidrule(lr){2-3}
        \cmidrule(lr){4-6}
        \cmidrule(lr){7-9}
        & Acc.
        & Pass
        & $\mathrm{mAP}_{50}$
        & $\mathrm{mAP}_{50:95}$
        & Recall
        & Cov.
        & Cat. prec.
        & Loc. prec. \\
        \midrule

        10
        & 73.11 & 98.71
        & 51.04 & 38.56 & 64.33
        & 41.69/13.20
        & 95.95/97.31
        & 96.06/44.44 \\

        20
        & 62.00 & 89.20
        & 59.38 & 44.62 & 67.31
        & 39.20/11.42
        & 97.22/99.42
        & 94.84/90.27 \\

        30
        & 54.43 & 74.93
        & 49.76 & 37.84 & 63.81
        & 29.07/9.63
        & 97.91/99.54
        & 96.53/70.62 \\

        40
        & 48.64 & 68.26
        & 50.43 & 38.88 & 70.29
        & 25.29/8.44
        & 97.36/99.21
        & 97.06/81.05 \\
        \bottomrule
    \end{tabular}
    \end{adjustbox}
    % \vspace{-8pt}
\end{table}

\subsection{Experimental Setup}
\label{subsec:exp_setup_metrics}

\begin{table*}[!t]
    \centering
    \caption{End-to-end VQA accuracy and computational efficiency of BrainFocus across Qwen3.5-VL model sizes.}
    \label{tab:end_to_end_results}

    \scriptsize
    \setlength{\tabcolsep}{3.1pt}
    \renewcommand{\arraystretch}{0.7}
    \newcommand{\BF}[1]{\textbf{#1}}

    \begin{adjustbox}{max width=\textwidth}
    \begin{tabular}{cccccccccccccc}
        \toprule

        \multirow{2}{*}{\textbf{Image Domain}}
        &
        \multirow{2}{*}{\textbf{Metric}}
        &
        \multicolumn{3}{c}{\textbf{10 classes}}
        &
        \multicolumn{3}{c}{\textbf{20 classes}}
        &
        \multicolumn{3}{c}{\textbf{30 classes}}
        &
        \multicolumn{3}{c}{\textbf{40 classes}}
        \\
        \cmidrule(lr){3-5}
        \cmidrule(lr){6-8}
        \cmidrule(lr){9-11}
        \cmidrule(lr){12-14}

        &
        &
        \textbf{2B} & \textbf{4B} & \textbf{9B}
        &
        \textbf{2B} & \textbf{4B} & \textbf{9B}
        &
        \textbf{2B} & \textbf{4B} & \textbf{9B}
        &
        \textbf{2B} & \textbf{4B} & \textbf{9B}
        \\

        \midrule

        % ==================== Generated cluttered ====================
        \multirow[c]{7}{*}{
            \shortstack[c]{
                \textbf{Generated cluttered image}\\[-1pt]
                \textit{Low target proportion}
            }
        }

        & Crop-detected acc. (\%)
        & 97.23 & 93.92 & 96.27
        & 91.84 & 94.33 & 93.82
        & 90.83 & 93.53 & 94.09
        & 91.52 & 94.29 & 93.76
        \\

        & Full-image acc. (\%)
        & 69.33 & 78.67 & 76.67
        & 73.00 & 79.00 & 76.00
        & 72.00 & 77.33 & 74.44
        & 71.83 & 78.00 & 74.50
        \\

        & \BF{BrainFocus acc. (\%)}
        & \BF{79.20} & \BF{84.62} & \BF{83.87}
        & \BF{80.42} & \BF{85.71} & \BF{84.69}
        & \BF{77.81} & \BF{82.68} & \BF{81.32}
        & \BF{77.07} & \BF{82.14} & \BF{79.59}
        \\

        & $\Delta$ acc. (pp)
        & +9.87 & +5.95 & +7.20
        & +7.42 & +6.71 & +8.69
        & +5.81 & +5.35 & +6.88
        & +5.24 & +4.14 & +5.09
        \\

        \cmidrule[0.20pt](lr){2-14}

        & Input-token red. (\%)
        & 39.4 & 39.4 & 39.4
        & 35.8 & 35.8 & 35.8
        & 26.5 & 26.5 & 26.5
        & 23.2 & 23.2 & 23.2
        \\

        & Total-token red. (\%)
        & 39.3 & 39.3 & 39.3
        & 35.8 & 35.7 & 35.8
        & 26.5 & 26.5 & 26.5
        & 23.2 & 23.2 & 23.2
        \\

        & FLOPs red. (\%)
        & 39.5 & 39.4 & 39.4
        & 35.9 & 35.8 & 35.8
        & 26.6 & 26.6 & 26.5
        & 23.3 & 23.3 & 23.2
        \\

        \midrule

        % ==================== Object-centric real ====================
        \multirow[c]{7}{*}{
            \shortstack[c]{
                \textbf{Real object-centric image}\\[-1pt]
                \textit{High target proportion}
            }
        }

        & Crop-detected acc. (\%)
        & 90.24 & 98.65 & 92.59
        & 92.61 & 92.61 & 91.25
        & 95.54 & 96.31 & 96.15
        & 89.61 & 93.55 & 91.84
        \\

        & Full-image acc. (\%)
        & 93.33 & 97.33 & 96.67
        & 92.33 & 96.00 & 93.67
        & 90.22 & 94.00 & 92.89
        & 89.67 & 94.00 & 93.00
        \\

        & \BF{BrainFocus acc. (\%)}
        & \BF{93.64} & \BF{97.73} & \BF{96.27}
        & \BF{92.84} & \BF{96.13} & \BF{94.00}
        & \BF{90.50} & \BF{93.90} & \BF{92.87}
        & \BF{89.83} & \BF{93.90} & \BF{92.88}
        \\

        & $\Delta$ acc. (pp)
        & +0.31 & +0.40 & $-0.40$
        & +0.51 & +0.13 & +0.33
        & +0.28 & $-0.10$ & $-0.02$
        & +0.16 & $-0.10$ & $-0.12$
        \\

        \cmidrule[0.20pt](lr){2-14}

        & Input-token red. (\%)
        & 6.8 & 6.8 & 6.8
        & 4.9 & 4.9 & 4.9
        & 3.5 & 3.5 & 3.5
        & 4.0 & 4.0 & 4.0
        \\

        & Total-token red. (\%)
        & 6.6 & 6.5 & 6.6
        & 4.7 & 4.7 & 4.8
        & 3.5 & 3.5 & 3.5
        & 4.0 & 4.0 & 4.0
        \\

        & FLOPs red. (\%)
        & 6.1 & 6.3 & 6.5
        & 4.1 & 4.4 & 4.7
        & 3.1 & 3.3 & 3.4
        & 3.7 & 3.8 & 3.9
        \\

        \bottomrule
    \end{tabular}
    \end{adjustbox}
\end{table*}

% 柱状图
\begin{figure*}[!t]
    \centering
    \includegraphics[width=1\textwidth]{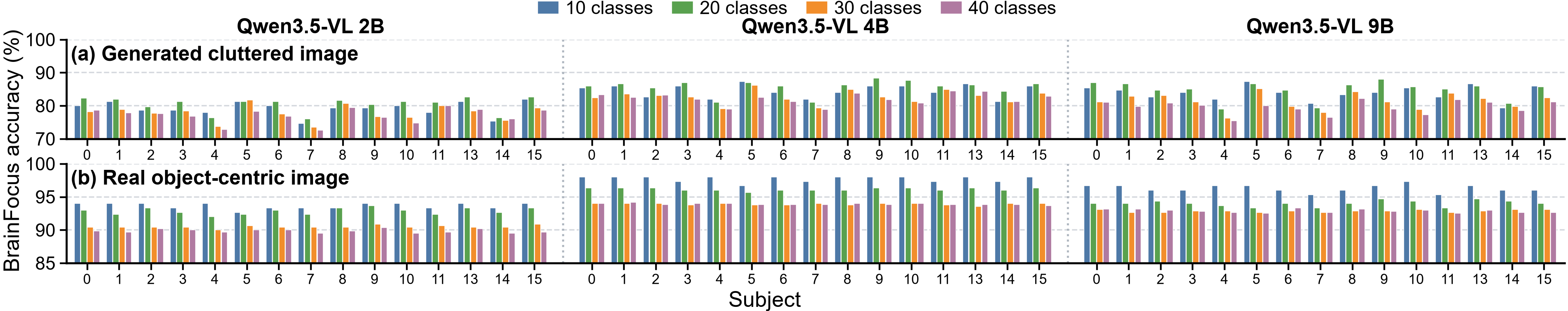}
    \caption{Subject-wise BrainFocus accuracy across 15 valid subjects, 4 class counts, 3 VLM model sizes (Qwen3.5-VL-2B, 4B and 9B), and 2 test domains.}
    \label{fig:subjectwise_framework}
    \vspace{-2mm}
\end{figure*}

We evaluate 10-, 20-, 30-, and 40-class settings over 15 subjects and both image domains. In stages 1 and 2, each category was trained, validated, and tested using 30/5/15 EEG trials or images, respectively. The generated images are used in all three stages, the real images are only used on the test set to obtain two different test domains.

We measure three types of accuracy. Full-image accuracy evaluates every request with the original full image. Crop-detected accuracy evaluates only accepted requests using their cropped ROIs. BrainFocus accuracy evaluates accepted requests with their crops and all remaining requests with the full-image fallback. We report $\Delta$ acc. as BrainFocus accuracy minus full-image accuracy in percentage points (pp). We recorded the reduction in input tokens and total tokens (input plus output), then reported the two reductions separately when using the full images as the input and when using our framework. We measured the FLOPs based on the actual inference calls of all three stages, and FLOPs reduction is the difference between the framework and the full-image VLM baseline.

\subsection{Stage-Wise Performance and Routing Behavior}

Table~\ref{tab:stagewise_results} reports the stage-wise test results using the confidence-gated strategy in Section~III-B. Acc. and Pass denote EEG top-1 accuracy and gate pass rate, while Cov. denotes the percentage of requests routed with a cropped ROI. Cat. prec. and Loc. prec. are the category and localization precision among accepted requests. We evaluate the stage 2 performance using mean average precision (mAP). $\mathrm{mAP}_{50}$ is computed at an intersection-over-union (IoU) threshold of 0.50, whereas $\mathrm{mAP}_{50:95}$ averages mAP over IoU thresholds from 0.50 to 0.95 in steps of 0.05. As the class count increases from 10 to 40, EEG accuracy decreases from 73.11\% to 48.64\%, and crop coverage decreases accordingly. Nevertheless, gating raises the accepted-category precision to 95.95--97.91\% on generated images and 97.31--99.54\% on real images. Localization precision remains above 94.84\% on generated images but is more variable on real images, indicating a synthetic-to-real domain gap. 
We next evaluate how this selective routing affects end-to-end VQA performance and computational efficiency.

\subsection{VQA Accuracy and End-to-End Efficiency}
\label{subsec:exp_end_to_end}

Table~\ref{tab:end_to_end_results} reports aggregate results, whereas Fig.~\ref{fig:subjectwise_framework} gives BrainFocus accuracy for each of the 15 subjects. On the generated-cluttered-image test domain, BrainFocus improves accuracy across all 12 VLM-size--class-count combinations by 4.14--9.87 pp, while reducing input tokens, total tokens, and FLOPs by 23.2--39.4\%, 23.2--39.3\%, and 23.2--39.5\%. On the real object-centric image test domain, accuracy changes by only $-0.40$ to $+0.51$ pp, while input tokens, total tokens, and FLOPs fall by 3.5--6.8\%, 3.5--6.6\%, and 3.1--6.5\% (Table~\ref{tab:end_to_end_results}).

For each VLM-size--class-count combination, subject-wise variation (highest subject accuracy minus lowest) on the generated-cluttered images is 5.67--9.00 pp. Thus, performance varies visibly across EEG subjects. The real object-centric images exhibit a tighter subject-wise distribution. The subject-wise variation is only 0.44--2.00 pp (Fig.~\ref{fig:subjectwise_framework}).

This contrast follows directly from the two image domains. Generated scenes contain more objects, more cluttered backgrounds, and smaller targets than the real object-centric images, making them closer to real-life scenario inputs. They are therefore harder, as reflected by their lower full-image accuracy (69.33--79.00\%, in Table~\ref{tab:end_to_end_results}) and their larger variation across EEG subjects. On the real images, the target is typically large and centered, and full-image accuracy is already high enough (89.67--97.33\%). BrainFocus therefore mainly preserves accuracy while reducing computation in this domain.

% crop-detected
Crop-detected accuracy reaches 90.83--97.23\% on the generated domain and 89.61--98.65\% on the real domain (Table~\ref{tab:end_to_end_results}), indicating that accepted ROIs support strong VQA performance in both domains. This metric is conditional on accepted requests. Future work should improve both Stages~1 and~2 to increase the non-fallback rate without compromising accuracy.

\section{Conclusion}
\label{sec:conclusions}

BrainFocus connects imperfect EEG decoding to model-agnostic VLM input reduction through confidence-gated ROI routing and full-image fallback. Across three Qwen3.5-VL sizes, it improves VQA accuracy by 4.14--9.87 pp on generated cluttered images while reducing input tokens, total tokens, and FLOPs by up to 39.4\%, 39.3\%, and 39.5\%. On real object-centric images, it preserves accuracy within $-0.40$ to $+0.51$ pp while reducing input tokens, total tokens, and FLOPs by up to 6.8\%, 6.6\%, and 6.5\%. BrainFocus ensures that in both cases, it largely preserves or improves the accuracy rate while achieving consistent computational savings. These findings support our key insight: EEG semantic decoding does not need to be correct on every trial to benefit multimodal inference when an appropriate strategy can reject unreliable cases.

\clearpage

\bibliographystyle{IEEEtran}
\bibliography{bibliography}

\end{document}